\documentclass{article}

\usepackage[final]{corl_2019} 

\usepackage{subfig}
\usepackage{graphicx}
\usepackage{booktabs}
\usepackage{amssymb}
\usepackage{multirow}
\usepackage{booktabs}
\usepackage{amsmath}
\usepackage{mathtools}
\usepackage{array}
\usepackage{combelow}
\usepackage{flushend}

\title{Robust Semi-Supervised Monocular Depth \\ Estimation with Reprojected Distances}

\author{Vitor Guizilini \quad Jie Li \quad Rare\cb{s} Ambru\cb{s} \quad Sudeep Pillai \quad Adrien Gaidon\\
Toyota Research Institute (TRI) \\
{\tt\ firstname.lastname@tri.global}}

\begin{document}
\maketitle


\begin{abstract}
Dense depth estimation from a single image is a key problem in computer vision, with exciting applications in a multitude of robotic tasks. Initially viewed as a direct regression problem, requiring annotated labels as supervision at training time, in the past few years a substantial amount of work has been done in self-supervised depth training based on strong geometric cues, both from stereo cameras and more recently from monocular video sequences.
In this paper we investigate how these two approaches (supervised \& self-supervised) can be effectively combined, so that a depth model can learn to encode true scale from sparse supervision while achieving high fidelity local accuracy by leveraging geometric cues. To this end, we propose a novel supervised loss term that complements the widely used photometric loss, and show how it can be used to train robust semi-supervised monocular depth estimation models.
Furthermore, we evaluate how much supervision is actually necessary to train accurate scale-aware monocular depth models, showing that with our proposed framework, very sparse LiDAR information, with as few as 4 beams (less than 100 valid depth values per image), is enough to achieve results competitive with the current state-of-the-art.


\end{abstract}

\keywords{Structure from Motion, Semi-Supervised Learning, Deep Learning, Depth Estimation, Computer Vision} 


\section{Introduction}

Depth perception is an essential component of any autonomous agent, enabling it to interact with objects and properly react to its surrounding environment. While there are sensors capable of providing such information directly, estimating depth from monocular images is particularly appealing, since cameras are inexpensive, compact, with low power consumption and capable of providing dense textured information. Recent breakthroughs in learning-based algorithms have allowed the generation of increasingly accurate monocular depth models, however these come with their own shortcomings. \textit{Supervised} methods \cite{fu2018deep} require additional sensors with precise cross-calibration to provide depth labels and normally do not generalize to non-supervised areas, while \textit{self-supervised} methods are limited by the accuracy of stereo \cite{pillai2018superdepth,zhou2018stereo,flynn2016deepstereo} or structure-from-motion \cite{garg2016unsupervised,zhou2017unsupervised,mahjourian2018unsupervised,monodepth17,GuiAl2019} reconstruction, with the latter also suffering from scale ambiguity. Nevertheless, \textit{supervised} methods still produce the highest accuracy models, especially when high quality groundtruth information is available, while \textit{self-supervised} methods are highly scalable, being capable of consuming massive amounts of unlabeled data to produce more generic models. Ideally, both approaches should be able to work together and complement each other in order to achieve the best possible solution, given the data that is available at training time.


Following this intuition, the main contribution of this paper is a novel supervised loss term that minimizes reprojected distances in the image space, 
and therefore operates under the same conditions as the 
photometric loss \cite{wang2004image}, which constitutes the basis for appearance-based self-supervised monocular depth learning methods. 
We show that this novel loss not only facilitates the injection of depth labels into self-supervised models, to produce \textit{scale-aware} estimates, but it also further improves the quality of these estimates, even in the presence of very sparse labels. The result is a novel semi-supervised training methodology that combines the best of both worlds, being able to consume massive amounts of unlabeled data, in the form of raw video sequences, while also properly exploiting the information contained in depth labels when they are available. The ability to properly leverage sparse information also greatly facilitates the generation of depth labels, eliminating the need for expensive 64 or 128-beam LiDAR sensors in favor of cheaper and easily available alternatives as the source of supervision.





 

\section{Related Work}


\subsection{Supervised Methods}

Supervised methods use ground truth depth, usually from LiDAR in outdoor scenes, to train a neural network as a regression model.
\citet{eigen2014depth} was amongst the first to propose convolutional neural networks as a solution to this problem, generating initially a coarse prediction and then refining it using another neural network to produce more accurate results. Since then, substantial work has been done to improve the accuracy of supervised depth estimation from monocular images, including the use of Conditional Random Fields (CRFs) \citep{depthcrf}, the inverse Huber distance loss function \citep{huberloss}, joint optimization of surface normals \citep{normalscvpr2}, fusion of multiple depth maps \citep{fouriercvpr} and more recently its formulation as an ordinal classification problem \citep{fu2018deep}. LiDAR data is also sparse relative to the camera field of view, and consequently supervised methods are unable to produce meaningful depth estimates in non-overlapping areas of the image.



\subsection{Self-Supervised Methods}

As supervised techniques for depth estimation advanced rapidly, the availability of target depth labels became challenging, especially for outdoor applications. To this end,~\cite{garg2016unsupervised,godard2017unsupervised} provided an alternative strategy involving training a monocular depth network with stereo cameras, without requiring ground-truth depth labels. By leveraging Spatial Transformer Networks~\cite{jaderberg2015spatial}, Godard et al~\cite{godard2017unsupervised} use stereo imagery to geometrically transform the right image plus a predicted depth of the left image into a synthesized left image. The loss between the resulting synthesized and original left images is then defined in a fully-differentiable manner, using a Structural Similarity~\cite{wang2004image} term and additional depth regularization terms, thus allowing the depth network to be self-supervised in an end-to-end fashion.
Following~\cite{godard2017unsupervised} and~\cite{ummenhofer2017demon}, Zhou et al.~\cite{zhou2017unsupervised} generalize this to self-supervised training in the \textit{purely} monocular setting, where a depth and pose network are simultaneously learned from unlabeled monocular videos. Several methods~\cite{yin2018geonet,mahjourian2018unsupervised,casser2018depth,zou2018dfnet,klodt2018supervising,wang2018learning, yang2018deep} have since then advanced this line of work by incorporating additional loss terms and constraints. 


\subsection{Semi-Supervised Methods}

Unlike both of the above mentioned categories, there has not been much work on semi-supervised depth estimation. Most notably, \citet{guo2018learning} and \citet{luo2018single} introduced multiple sequential self-supervised and supervised training stages, where each stage is conducted independently. \citet{kuznietsov2017semi} proposed adding the supervised and self-supervised loss terms together, allowing both sources of information to be used simultaneously. The same concept was applied in \cite{SemiLeftRight}, with the introduction of left-right consistency to avoid post-processing at inference time. Our work follows a similar direction, focusing on how to properly incorporate depth labels into appearance-based self-supervised learning algorithms at training time, so we can both produce scale-aware models and further improve on the quality of depth estimates, even in the presence of very sparse labels.

\section{Methodology}

Our proposed semi-supervised learning methodology is composed of two training stages, as depicted in Figure 1. The first consists of a self-supervised monocular structure-from-motion setting, where we aim to learn: (i) a monocular depth model $f_D: I \to D$, that predicts the scale-ambiguous depth $\hat{D} = f_D(I(p_t))$ for every pixel $p_t$ in the target image $I_t$; and (ii) a monocular ego-motion model $f_{\mathbf{x}}: (I_t,I_S) \to \mathbf{x}_{t \to S}$, that predicts the set of 6-DoF rigid transformations for all $s \in S$ given by $\mathbf{x}_{t \to s} = \begin{psmallmatrix}\mathbf{R} & \mathbf{t}\\ \mathbf{0} & \mathbf{1}\end{psmallmatrix} \in \text{SE(3)}$, between the target image $I_t$ and the set of source images $I_s \in \mathcal{S}$ considered as part of the temporal context. Afterwards, if supervision is available, this information can be used to (i) collapse the scale ambiguity inherent to a single camera configuration into a metrically accurate model, and (ii) further improve the depth and ego-motion models by leveraging cues that are not appearance-based. More details on network architectures and objective function terms are given in the following sections.



\subsection{Depth Network}

Our depth network is based on the architecture 
introduced by \citet{GuiAl2019}, that proposes novel packing and unpacking blocks to respectively downsample and upsample feature maps during the encoding and decoding stages (Fig. \ref{fig:diagram}). Skip connections~\cite{mayer2016large} are used to facilitate the flow of information and gradients throughout the network. The decoder produces intermediate inverse depth maps, that are upsampled before being concatenated with their corresponding skip connection and unpacked feature maps. They also serve as output to produce a 4-scale inverse depth pyramid, from which the loss is calculated. However, instead of incrementally super-resolving each inverse depth map, as described in  \citep{GuiAl2019}, here they are all upsampled directly to the highest resolution using bilinear interpolation. As noted in~ \citep{godard2018digging2}, this reduces copy-based artifacts and photometric ambiguity, leading to better results as shown in experiments.

\subsection{Pose Network}

\begin{figure}[t]
    \centering
    \includegraphics[width=0.95\linewidth]{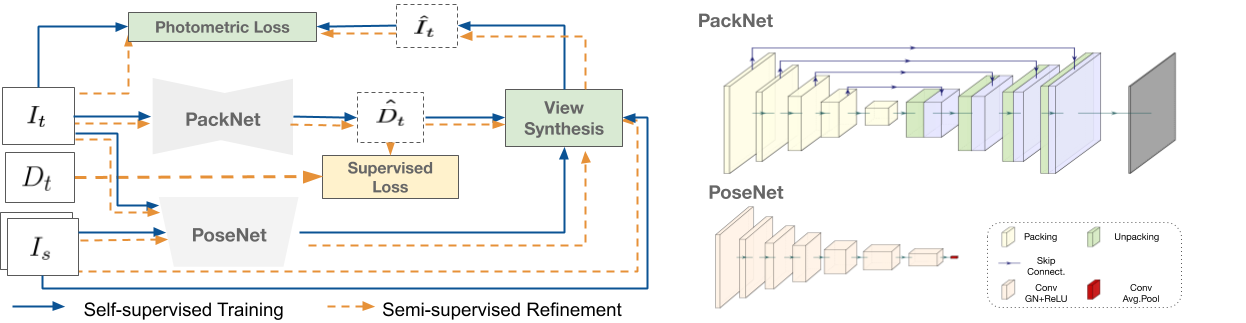}
    \caption{\textbf{Diagram of the proposed semi-supervised} monocular depth estimation framework.}
    \label{fig:diagram}
\end{figure}

For camera ego-motion estimation, we use the neural network architecture proposed by~\cite{zhou2017unsupervised} \textit{without} the explainability mask, which we found not to improve results. Following~\cite{zhou2017unsupervised}, the pose network consists of 7 convolutional layers followed by a final $1 \times 1$ convolutional layer and a 6-channel average pooling operation (Fig. \ref{fig:diagram}). The input to the network consists of a target $I_t$ and context $I_s$ images, concatenated together, and the output is the set of 6 DoF transformations between $I_t$ and $I_s$ that constitute $T_{s \rightarrow t}$. If more than one contextual image is considered, this process is repeated for each $I_s \in \mathcal{S}$ to produce independent transformations. Translation is parametrized in Euclidean coordinates $\{x, y, z\}$ and rotation uses Euler angles $\{\alpha, \beta, \gamma\}$.

\subsection{Objective Function}

The objective function used in our work has two components: a \textit{self-supervised} term, that operates on appearance matching $\mathcal{L}_{photo}$ between the target $I_t$ and synthesized images $I_{s \rightarrow t}$ from the context set $\mathcal{S} = \{I_s\}_{s=1}^S$, with masking $M_{photo}$ and depth smoothness $\mathcal{L}_{smooth}$; and a \textit{supervised} term, that operates on the reprojected $\mathcal{L}_{rep}$ distances between predicted and ground-truth depth values. The coefficients $\lambda$ are responsible for weighting the various terms relative to the photometric loss. The full objective function is as follows:
\begin{align}
\mathcal{L}(I_t,\mathcal{S}) &= \mathcal{L}_{photo} \odot M_{photo} + \lambda_{smooth} \cdot  \mathcal{L}_{smooth}  +
\lambda_{rep} \cdot \mathcal{L}_{rep}
\label{eq:overall-loss}
\end{align}
\textbf{Appearance Matching Loss.}~~Following~\cite{zhou2017unsupervised} the pixel-level similarity between the target image $I_t$ and the synthesized target image $I_{t \rightarrow s}$ is estimated using the Structural Similarity (SSIM)~\cite{wang2004image} term combined with an L1 pixel-wise loss term, inducing an overall photometric loss given by:
\begin{equation}
\mathcal{L}_{photo}(I_t,I_{s \rightarrow t}) = \alpha~\frac{1 - \text{SSIM}(I_t,I_{s \rightarrow t})}{2} + (1-\alpha)~\| I_t - I_{s \rightarrow t} \|
\label{eq:loss-photo}
\end{equation}
While multi-view projective geometry provides strong cues for self-supervision, errors due to parallax and out-of-bounds objects have an undesirable effect incurred on the photometric loss, that adds noise to the training stage and should not be learned by the depth model. Following \citet{godard2018digging2}, we mitigate these undesirable effects by calculating the minimum photometric loss per pixel for each source image in the context $\mathcal{S}$, so that:
\begin{equation}
\mathcal{L}_{photo}(I_t, \mathcal{S}) = \min_{s \in \mathcal{S}} \mathcal{L}_{photo}(I_t,I_{s\rightarrow t})
\end{equation}
The intuition is that the same pixel will not be occluded or out-of-bounds in all context images, and that the association with minimal photometric loss should be the correct one. Furthermore, also following \citep{godard2018digging2} we mask out static pixels by removing those which have a \textit{warped} photometric loss $\mathcal{L}_{photo}(I_t, I_{t \rightarrow s})$ higher than their corresponding \textit{unwarped} photometric loss $\mathcal{L}_{photo}(I_t, I_{s})$, calculated using the original source image without view synthesis. This mask removes pixels whose appearance does not change between frames, which includes static scenes and dynamic objects moving at a similar speed as the camera, since these will have a smaller photometric loss when we assume no ego-motion.
\begin{equation}
M_{photo} = \min_{s \in \mathcal{S}} \mathcal{L}_{photo}(I_t, I_s) > \min_{s \in \mathcal{S}} \mathcal{L}_{photo}(I_t, I_{s \rightarrow t})
\end{equation}

\textbf{Depth Smoothness Loss.}~~In order to regularize the depth in texture-less low-image gradient regions, we incorporate an edge-aware term, similar to~\cite{godard2017unsupervised}. The loss is weighted for each of the pyramid levels, starting from 1 and decaying by a factor of 2 on each scale:
\begin{equation}
\mathcal{L}_{smooth}(\hat{D}_t) = | \delta_x \hat{D}_t | e^{-|\delta_x I_t|} + | \delta_y \hat{D}_t | e^{-|\delta_y I_t|}
  \label{eq:loss-disp-smoothness}
\end{equation}

\begin{figure}
\centering
\begin{tabular}{cc}
\multirow{-8}[2]{*}{\subfloat[Reprojected distance loss]{\includegraphics[width=0.43\textwidth]{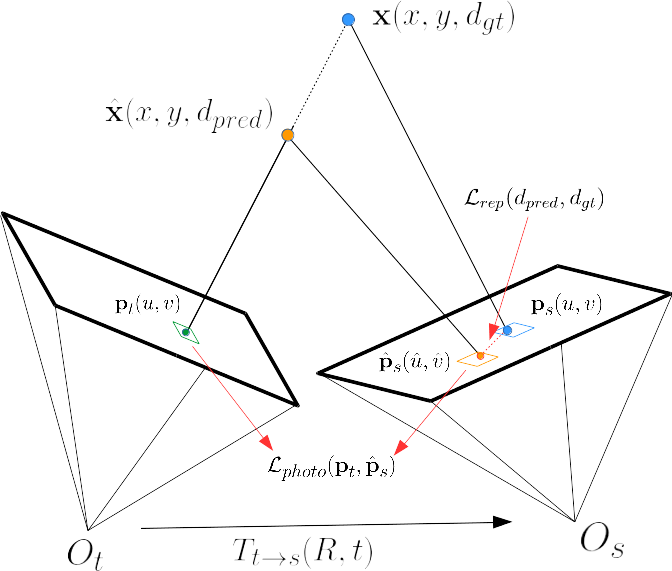}
\label{fig:2dloss}}} \hspace{0.3cm}
&
\subfloat[Loss distribution (real sample)]{\includegraphics[width=0.35\textwidth]{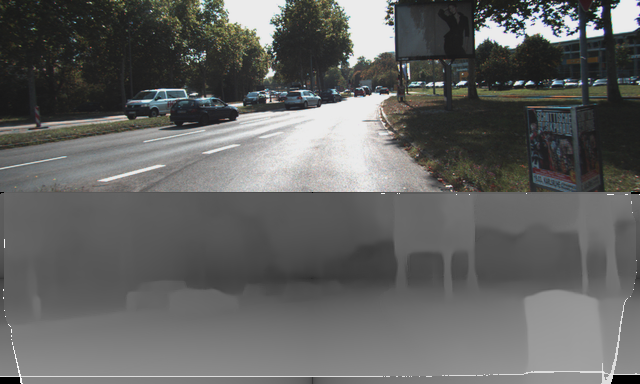}
\label{fig:errordist1}}  
\\
& \subfloat[Loss distribution over uniform error]{\includegraphics[width=0.35\textwidth]{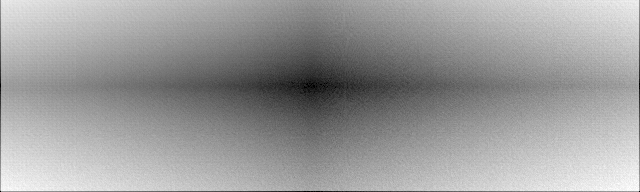}
\label{fig:errordist2}} 
\\
\end{tabular}
\caption{\textbf{Diagram of the proposed reprojected distance loss} and its behavior relative to $\textbf{p}_t$ pixel coordinates. In (b), the bottom row depicts the per-pixel reprojected loss values when there is a constant error of $1$m relative to the predicted depth values from the top row image. Similarly, (c) depicts the per-pixel loss values when the same $1$m error is applied to a constant depth of $50$m throughout the entire image.}
\vspace{-0.4cm}
\end{figure}

\textbf{Reprojected Distance Loss.}
\label{sec:reploss}~~Most supervised depth training methods operate on a direct regression basis, mapping input RGB images into output depth values without exploiting motion information or camera geometry. However, if such information is available \textemdash~as is the case in self-supervised learning \textemdash, we show here that it can be leveraged to produce more accurate and consistent depth models. This is achieved using a novel supervised loss term (Eq. \ref{eq:reploss}) that operates under the same conditions as the photometric loss, by reprojecting depth errors back onto the image space as observed by a context camera, as shown in Fig. \ref{fig:2dloss}. For each pixel $\textbf{p}_t^i$ in the target image, this reprojected depth error $\mathcal{L}_{rep}$ corresponds to the distance between its true $\textbf{p}_s^i$ and predicted $\hat{\textbf{p}}_s^i$ associations in the source image, from which the photometric loss $\mathcal{L}_{photo}$ is calculated: 
\begin{align}
\mathcal{L}_{rep}(\hat{D}_t, D_t) =& \frac{1}{V} \sum_{i \in \mathcal{V}_t}
\| 
\hat{\textbf{p}}_s^i - \textbf{p}_s^i 
\| = 
\frac{1}{V} \sum_{i \in \mathcal{V}_t}
\| 
\mathcal{\pi}_s( \hat{\textbf{x}}_t^i)- 
\mathcal{\pi}_s( {\textbf{x}}_t^i) 
\| 
\nonumber \\
=& \frac{1}{V} \sum_{i \in \mathcal{V}_t}
\|
\mathcal{\pi}_s( \hat{d}_t^i K^{-1}\textbf{u}_t^i)- 
\mathcal{\pi}_s( {d}_t^i K^{-1}\textbf{u}_t^i)
\|
\label{eq:reploss}
\end{align}
where $\textbf{u}_t^i = (u,v,1)_s^{i, T}$ denotes the homogeneous coordinates of pixel $i$ in target image $I_t$, and $\hat{\textbf{x}}_t^i$ and ${\textbf{x}}_t^i$ are the homogeneous coordinates of its reconstructed 3D points given respectively the predicted $\hat{d}^i$ and ground truth $d^i$ depths values. $\hat{\textbf{p}}^i_s=(\hat{u},\hat{v})_s^{i,T}$ and $\textbf{p}^i_s=(u,v)_s^{i,T}$ denote respectively the 2D projected pixel coordinates of points $\hat{\textbf{x}}_t^i$ and ${\textbf{x}}_t^i$ onto source frame $I_s$, produced by the project function $\mathcal{\pi}_t=\mathcal{\pi}(\textbf{x} ;K,T_{t\rightarrow s})$. The effects of operating in this reprojected space are shown in Figs. \ref{fig:errordist1} and \ref{fig:errordist2}, where we respectively see how depth ranges are weighted differently, and so are pixel coordinates closer to the vanishing point. The concept of depth weighting has been explored before, such as with the inverse Huber loss \citep{huberloss} and spacing-increasing discretization \citep{fu2018deep}, however here
this weighting is not artificially introduced, but rather comes as a direct by-product  of camera geometry, with errors being proportional to their respective reprojections onto the image space. Similarly, weighting based on distance to the vanishing point directly correlates to the baseline used in structure-from-motion calculation. If there is no relative baseline (i.e. it coincides with the error vector $\textbf{e}_t^i = \textbf{x}_t^i - \hat{\textbf{x}}_t^i$), this is an ill-defined problem, and therefore no depth information can be produced by that particular configuration. 

Note that this proposed loss term is not appearance-based, and therefore in theory any transformation matrix $T$ could be used to produce the reprojections from which the distance is minimized. However, by enforcing $T = T_{s \rightarrow t}$ we can (a) back-propagate through the pose network, so it can also be directly updated with label information and remain consistent with the depth network; and (b) operate on the same reprojected distances that are used by the photometric loss, only on a \textit{scale-aware} capacity, so its inherent ambiguity is forced to collapse into a metrically accurate model.


\section{Experimental Results}

\subsection{Datasets}

We use the popular KITTI~\cite{geiger2013vision} dataset for all experiments, to facilitate comparisons with other related methods. The evaluation method introduced by Eigen \cite{eigen2014depth} for KITTI uses raw reprojected LIDAR points, 
and does not handle occlusions, dynamic objects or ego-motion, which leads to wrong image reprojections.  A new set of high quality depth maps for KITTI was introduced in \cite{gtkitti}, making use of 5 consecutive frames to increase the number of available information and handling moving objects using the stereo pair.  Throughout our experiments, we refer to this new set as \textit{Annotated}, while the original set is referred to as \textit{Original}. For training, we use the pre-processing methodology described in \cite{zhou2017unsupervised} to remove static frames, resulting in $39810$ training images from both left and right cameras (note that no stereo information is used in this work) and $652$ evaluation images. Context information for monocular depth and pose learning is generated using the immediate previous $t-1$ and posterior $t+1$ frames.



\subsection{Implementation Details}

We use PyTorch~\cite{paszke2017automatic} for all our experiments\footnote{Inference code and pretrained weights for our models are available upon request.}, training across 8 distributed Titan V100 GPUs. We use the Adam optimizer~\cite{kingma2014adam}, with $\beta_1=0.9$ and $\beta_2=0.999$. When training a model from scratch, we follow \citet{GuiAl2019} and optimize the depth and pose networks for $100$ epochs, with a batch size of 4 and initial depth and pose learning rates of $2\cdot10^{-4}$, that are halved every 40 epochs. We set the SSIM weight to $\alpha=0.85$ (Eq. \ref{eq:loss-photo}), the depth smoothness weight to $\lambda_{smooth}=10^{-3}$ (Eq. \ref{eq:overall-loss}) and, when applicable, the reprojected distance weight to $\mathcal{L}_{rep}=10^4$ (Eq. \ref{eq:overall-loss}). When refining a model only $15$ epochs are considered, with depth and pose learning rates of $10^{-4}$ that are halved every $6$ epochs, and the same weight values for the loss function are used. Input images are downsampled to a $640 \times 192$ resolution, and horizontal flipping and jittering are used as data augmentation.


\subsection{Depth Estimation}

\begin{figure}[b!]
    \vspace{-0.3cm}
	\centering
	\subfloat[Semi-Supervision with Reprojected Distance loss]{
		\includegraphics[width=0.49\textwidth]{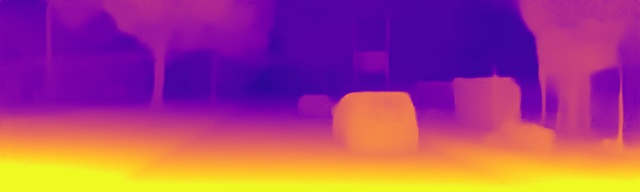}} 
	\subfloat[Semi-Supervision with L1 loss ]{
		\includegraphics[width=0.49\textwidth]{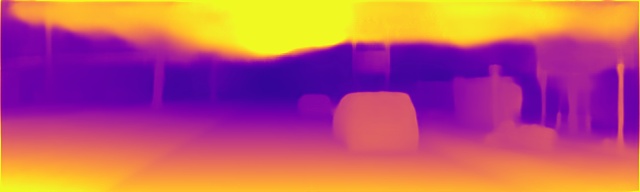}}
	\\ \vspace{-3mm}
	\subfloat[Semi-Supervision with BerHu loss]{
		\includegraphics[width=0.49\textwidth]{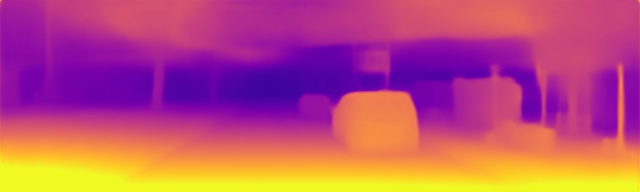}}
\subfloat[DORN \cite{fu2018deep} results]{
        \label{fig:dorn}
		\includegraphics[width=0.49\textwidth]{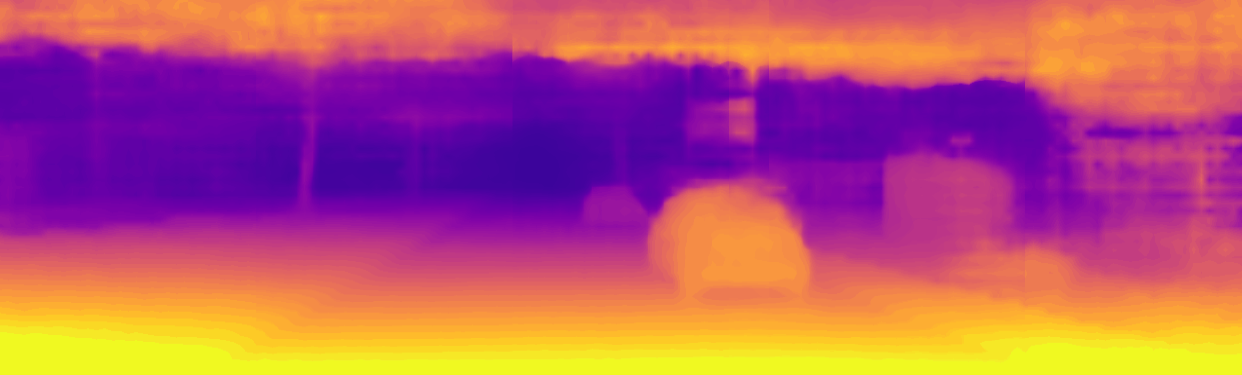}}		
	\caption{\textbf{Example of depth maps} produced by an unscaled self-supervised model, and after its refinement with different supervised losses (the colormaps used for plotting are produced to be scale agnostic). As expected, our proposed Reprojected Distance loss is able to better reconstruct areas not observed by the LiDAR sensor used as source of supervision. For comparison, we also show results using the current state-of-the-art supervised method from \cite{fu2018deep}, which also fails to properly reconstruct areas not observed by the LiDAR sensor at training time.} 
\label{fig:skylearning}
\end{figure}

In this section we present and discuss our results in monocular depth estimation, and how they compare with similar works found in the literature. An ablative study of the different components of our proposed semi-supervised training methodology can be found in Table \ref{tab:ablation}, starting with quantitative evidence that our proposed modifications to the losses used by the original PackNet-SfM framework \cite{GuiAl2019} are able to improve self-supervised depth training, resulting in significantly more accurate models. These results are compared against other techniques in Table \ref{table:depth-accuracy} (\textit{Original} depth maps), and constitute a new state-of-the-art in self-supervised depth learning, surpassing even stereo methods. Given this high-quality unscaled model, we proceed to show how depth labels can be efficiently incorporated at training time to further improve results. Initially, we train two supervised models with the traditionally used L1 loss (absolute depth distance), both starting from scratch and from our pretrained self-supervised checkpoint. We then introduce the semi-supervision methodology described in this paper, by using both the photometric and L1 losses during the refinement process from the same pretrained self-supervised checkpoint. Finally, we remove the L1 loss and introduce our proposed reprojected distance loss, training again in a semi-supervised manner two models, one from scratch and another from the pretrained self-supervised checkpoint.

\begin{table*}[!t]
	\centering
	{
		\small
		\setlength{\tabcolsep}{0.3em}
		\begin{tabular}{lc|c|c|c|ccccc}
			\toprule
			\textbf{Method} &
			Supervision & 
			MS & RF & Loss &
			Abs.Rel &
			Sq.Rel &
			RMSE &
			RMSE$_{log}$ &
			$\delta < 1.25$
			\vspace{0.5mm}\\
			\toprule

			
PackNet-SfM \citep{GuiAl2019} & Self & \checkmark & & & 0.086 & 0.460 & 3.712 & 0.132 & 0.918 \\			

\textbf{Ours (baseline)} & Self & \checkmark & & & 0.078 & 0.417 & 3.487 & 0.121 & 0.931 \\

\textbf{Ours} & Semi & \checkmark & \checkmark & Rep & \textbf{0.069} & \textbf{0.313} & \textbf{3.097} & \textbf{0.110} & \textbf{0.941} \\

\midrule

\textbf{Ours} & Sup & & \checkmark & L1 & 0.078 & 0.378 & 3.330 & 0.121 & 0.927 \\
\textbf{Ours} & Semi & &  & L1 & 0.084 &0.437 & 3.68 & 0.133 & 0.913\\
\textbf{Ours} & Semi & & \checkmark & L1 & 0.078 & 0.383 & 3.340 & 0.121 & 0.927\\

\textbf{Ours} & Semi & & & Rep & 0.077 & 0.362 & 3.274 & 0.119 & 0.929 \\

\textbf{Ours} & Semi & & \checkmark &  Rep & \textbf{0.072} & \textbf{0.340} & \textbf{3.265} & \textbf{0.116} & \textbf{0.934} \\

\bottomrule

\end{tabular}
	}\vspace{1mm}
	\caption{\textbf{Ablation study of our proposed framework}, on the KITTI dataset with $640 \times 192$ input resolution and \textit{Annotated} depth maps when applicable. $MS$ indicates the use of ground-truth median-scaling at inference time. \textit{Self}, \textit{Sup} and \textit{Semi} respectively indicate self-supervised, supervised and semi-supervised training. $RF$ indicates the use of \textbf{Ours (baseline)} as a pretrained model for refinement. $L1$ and $Rep$ respectively indicate supervision using the L1 loss (absolute depth distance) and the proposed reprojected distance loss	(Sec. \ref{sec:reploss}).}

	
	
	\label{tab:ablation}
\vspace{-5mm}
\end{table*}

Interestingly, we can see from these results that the introduction of depth labels for semi-supervision, when using the L1 loss, mostly enabled the model to learn scale, however it was unable to further improve the overall quality of depth estimates (i.e. going from $0.078$ with median-scaling to $0.078$ without median-scaling). In contrast, semi-supervision with our proposed reprojected distance loss effectively improved the accuracy of the refined depth model, going from $0.078$ with median-scaling to $0.072$ without median-scaling. Furthermore, we show in Fig. \ref{fig:skylearning} that our proposed loss term also enables the learning of better estimates in areas where there is no LiDAR supervision, such as the upper portions of the image. We attribute this behavior to the fact that the reprojected distance loss operates in the same image space as the photometric loss, and thus is better suited to address its inherent scale ambiguity, consistently collapsing the entire image to a metrically accurate model. We also compare these results with other similar techniques  found in the literature (Table \ref{table:depth-accuracy}, \textit{Annotated} depth maps), and show that they are competitive with the current state-of-the-art, surpassing all other semi-supervised methods and achieving similar performance\footnote{A video with further results can be found in \url{https://www.youtube.com/watch?v=cSwuF-XA4sg}} as the Deep Ordinal Regression Networks proposed in \cite{fu2018deep}. However, because our proposed approach also leverages unlabeled data via the photometric loss, we are able to process the entire image at training time, thus producing more visually consistent depth maps, as shown in Fig. \ref{fig:dorn}. Note that our pose network also learns to produce metrically-accurate estimates ($t_{rel}$ and $r_{rel}$ of $2.4296$ and $0.5747$ respectively on the KITTI odometry benchmark for training sequences $00/03/04/05/07$, and $6.8017$ and $2.7142$ for testing sequences $01/02/06/08/10$), even though there was no direct pose supervision.




\begin{table*}[t!]
	\centering
	{
		\small
		\setlength{\tabcolsep}{0.3em}
		\begin{tabular}{c|lccccccccccccc}
			\toprule
			&
			\textbf{Method} &
			Supervision & 
			Resolution & 
			Dataset &
			Abs.Rel &
			Sq.Rel &
			RMSE &
			RMSE$_{log}$ &
			$\delta < 1.25$ \\
			\midrule
			
\parbox[t]{2mm}{\multirow{6}{*}{\rotatebox[origin=c]{90}{Original}}}

& Monodepth2.~\cite{godard2018digging2}$^\ddagger$ 
& Self (M) & 640 x 192 & K & 0.129 & 1.112 & 5.180 & 0.205 & 0.851 \\

& PackNet-SfM \citep{GuiAl2019}
& Self (M) & 640 x 192 & K & 0.120 & 1.018 & 5.136 & 0.198 & 0.865 \\

& Monodepth2~\cite{godard2018digging2}$^\ddagger$  
& Self (S) &  640 x 192 &  K & 0.115 & 1.010 & 5.164 & 0.212 & 0.858 \\


			
& Monodepth2 \citep{godard2018digging2}$^\ddagger$
& Self (M) & 640 x 192 & K & 0.115 & 0.903 & 4.863 & 0.193 & 0.877 \\

& SuperDepth~\cite{pillai2018superdepth} & Self (S) & 1024 x 384 & K & 0.112 & 0.875 & 4.958 & 0.207 & 0.852 \\


& \textbf{Ours} 
& Self (M) & 640 x 192 & K & \textbf{0.111} & \textbf{0.785} & \textbf{4.601} & \textbf{0.189} & \textbf{0.878} \\



\midrule

\parbox[t]{2mm}{\multirow{11}{*}{\rotatebox[origin=c]{90}{\small Annotated}}}		

			

& \citet{kuznietsov2017semi}$^\ddagger$
& Semi (S) & 640 x 192 & K & 0.089 & 0.478 & 3.610 & 0.138 & 0.906 \\

& SemiDepth \citep{SemiLeftRight} 
& Semi (S) & 640 x 192 & C+K & 0.078 & 0.417 & 3.464 & 0.126 & 0.923 \\

& SVSM FT \citep{luo2018single} $^\ddagger$
& Semi (S) & 640 x 192 & F+K & 0.077 & 0.392 & 3.569 & 0.127 & 0.919 \\

& DORN (VGG) \citep{fu2018deep} $^\ddagger$ 
& Sup & 640 x 192 & K & 0.081 & 0.376 & 3.056 & 0.132 & 0.915 \\

& DORN (ResNet) \citep{fu2018deep}$^\ddagger$ 
& Sup & 640 x 192 & K & \textbf{0.072} & \textbf{0.307} & \textbf{2.727} & 0.120 & 0.930 \\


& \textbf{Ours} 
& Semi (M) & 640 x 192 & K & \textbf{0.072} & 0.340 & 3.265 & \textbf{0.116} & \textbf{0.934} \\

\cmidrule{2-10} 

& \textbf{Ours~~(64 beams)}
& Semi (M) & 640 x 192 & K & 0.074 & 0.355 & 3.349 & 0.118 & 0.930 \\

& \textbf{Ours~~(32 beams)} 
& Semi (M) & 640 x 192 & K & 0.076 & 0.363 & 3.361 & 0.121 & 0.929 \\

& \textbf{Ours~~(16 beams)} 
& Semi (M) & 640 x 192 & K & 0.078 & 0.371 & 3.388 & 0.122 & 0.928 \\

& \textbf{Ours~~~~(8 beams)} 
& Semi (M) & 640 x 192 & K & 0.078 & 0.395 & 3.587 & 0.126 & 0.922 \\

& \textbf{Ours~~~~(4 beams)} 
& Semi (M) & 640 x 192 & K & 0.082 & 0.424 & 3.732 & 0.131 & 0.917 \\

\bottomrule

		\end{tabular}
	}\vspace{1mm}
\caption{\textbf{Quantitative comparison} between different learning-based monocular depth estimation techniques. \textit{Self}, \textit{Sup} and \textit{Semi} respectively indicate self-supervised, supervised and semi-supervised training, with (M) and (S) respectively indicating monocular and stereo self-supervision. $F$, $C$ and $K$ indicate respectively the FlyingThings3D, Cityscapes and KITTI datasets. $\ddagger$ indicates ImageNet \citep{Deng09imagenet} pretraining on the depth and/or pose networks. All \textit{Mono} methods use median-scaling at inference time, to produce metrically accurate estimates.}
	
	
	
	\label{table:depth-accuracy}
	\vspace{-2mm}
\end{table*}

\begin{figure}[b!]
    \vspace{-0.5cm}
	\centering
	\subfloat[\textit{Annotated} depths ($18288$ points)]{ \includegraphics[width=0.33\textwidth]{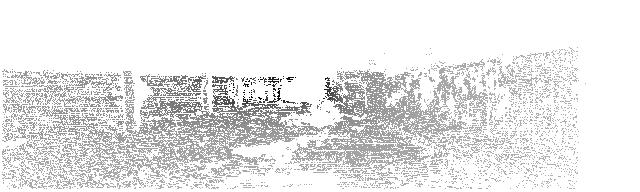}}	
	\subfloat[64 beams ($1427$ points)]{
	\includegraphics[width=0.33\textwidth]{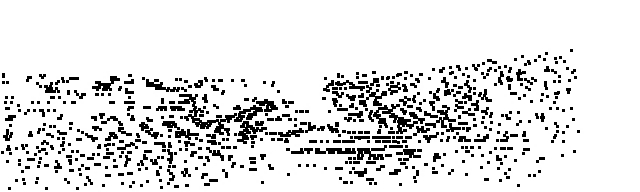}} 
	\subfloat[32 beams (711 points)]{
	\includegraphics[width=0.33\textwidth]{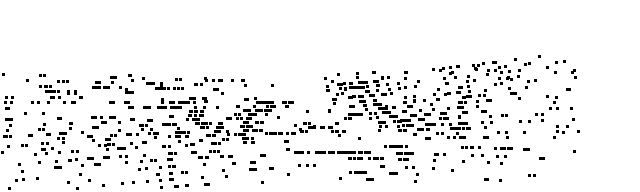}} 
	\\ \vspace{-0.3cm}
	\subfloat[16 beams (347 points)]{
	\includegraphics[width=0.33\textwidth]{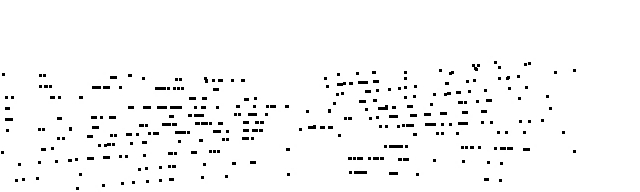}} 
	\subfloat[8 beams (171 points)]{ \includegraphics[width=0.33\textwidth]{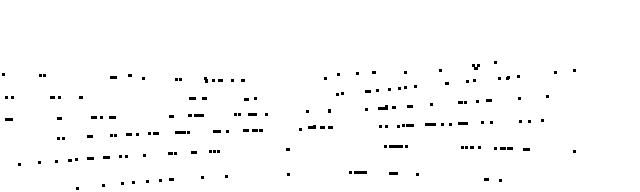}} 
	\subfloat[4 beams (77 points)]{
	\includegraphics[width=0.33\textwidth]{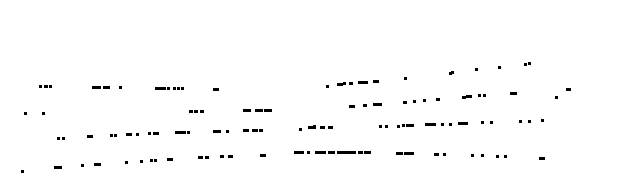}}   
	\caption{\textbf{Effects of LiDAR beam decimation} on the sparsity of depth labels. Relative to a full annotated depth map, a 4-beam representation contains only $0.42\%$ of the number of valid depth values, or $0.06\%$ of the total number of pixels in the image.}
	\label{fig:depthmaps}
\end{figure}

\begin{figure}[h]
	\centering
	\subfloat[Absolute Relative Error ($abs\_rel$).]{
		\includegraphics[width=0.50\textwidth, height=0.25\textwidth]{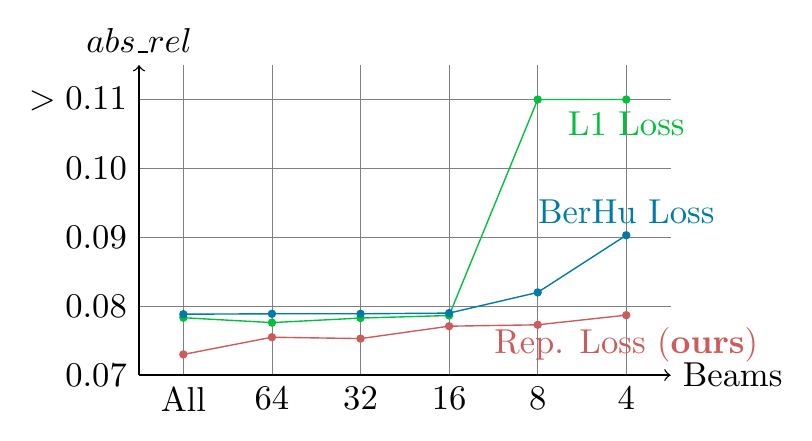}}
	\subfloat[Threshold 1 ($\delta < 1.25$)]{
		\includegraphics[width=0.50\textwidth,height=0.25\textwidth]{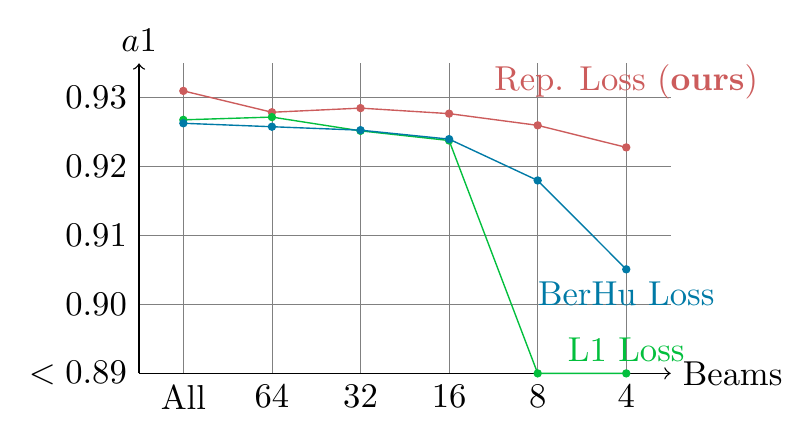}} \\
	\caption{\textbf{Depth degradation with a decreasing number} of LiDAR beams, for semi-supervised training with different losses. Loss weight for each type of loss are optimized through grid search.}
    \label{fig:decimation_chart}
\end{figure}

\subsection{Sparse Depth Labels}


The \textit{Annotated} depth maps, used in this work and in most related works on supervised depth learning, have been carefully curated and are composed of accumulated LiDAR beams, resulting in information that is much denser than what is produced by traditional range sensors. In fact, sparse LiDAR sensors are not uncommon in most real-world robotics platforms, and devising ways to properly leverage this information for learning-based methods would be highly valuable as a way to decrease costs and increase data collection rate. In this section we investigate the effects that sparse information has in supervised depth training, and how our proposed semi-supervised training methodology is able to mitigate degradation in depth estimates when labels are substantially sparser than commonly reported in related works. This sparsity effect can be achieved by selectively masking out pixels from each of the $64$ LiDAR beams, keeping only those belonging to equally spaced beams at increasing intervals. To maintain consistency with previous experiments, these masks are produced individually for each frame and the \textit{original} depth values at each valid pixel is substituted by the corresponding \textit{annotated} values. The effects of decimating depth maps in such a way is shown in Fig. \ref{fig:depthmaps}, where we can see a substantial decrease in the amount of information available for supervised training as more beams are removed, reaching less than 100 valid pixels per image when only four beams are considered. 

Even so, in Fig. \ref{fig:decimation_chart} we can see that our proposed approach is capable of producing accurate depth estimates even when using as few as 4 beams for depth supervision at training time. In contrast, semi-supervision with other traditional losses starts to degrade after reaching a certain level of sparsity, empirically determined to be at $16$ beams, or around $2\%$ of the original number of valid pixels. This behavior is more visible on the L1 loss (green line), while the BerHu loss (blue line) degrades at a slower pace, however it still starts to decay exponentially, while our proposed Reprojected Distance loss (red line) maintains a roughly linear decay relative to the number of considered beams. This decay is also numerically represented in Table \ref{table:depth-accuracy}, for all considered metrics. In fact, our results when using only 8 beams are comparable to \citet{SemiLeftRight} and \citet{luo2018single}, considered the current state-of-the-art for semi-supervised monocular depth estimation, and our results when using only 4 beams are better than \citet{kuznietsov2017semi}.



\section{Conclusion}

This paper introduces a novel semi-supervised training methodology for monocular depth estimation, that both improves on the current state of the art and is also more robust to the presence of sparse labels. To accomplish this, we propose a new supervised loss term that operates in the image space, and thus is compatible with the widely used photometric loss in the semi-supervised setting.
We show, using the popular KITTI benchmark, that our proposed methodology can efficiently incorporate information from depth labels into pretrained self-supervised models, allowing them to produce metrically accurate estimates while further improving the overall quality of resulting depth maps. Further analysis also indicates that our model presents a strong robustness to the degradation of available supervised labels, reaching results competitive with the current state-of-the-art even when as few as 4 beams are considered at training time. Future work will focus on the scalability of the proposed semi-supervised training methodology, investigating its application to different datasets, sources of labeled information and robustness to domain transfer.


\clearpage
\appendix



\section{Self-Supervised and Supervised Losses Trade-Off}

Our proposed semi-supervised loss (Eq. 1, main text) is composed of three individual terms: $\mathcal{L}_{photo}$, representing the self-supervised photometric loss, $\mathcal{L}_{smooth}$, representing a self-supervised smoothness depth regularizer, and $\mathcal{L}_{rep}$, representing the proposed supervised reprojected distance loss. Determining the correct balance between these terms is an important part of the training protocol, and in this section we discuss the effects that $\lambda_{rep}$, or the ratio between the self-supervised and supervised components of the loss, has in our overall results.

\begin{figure*}[b]
	\centering
	\subfloat[$\lambda_{rep} = 10^4$]{
		\includegraphics[width=0.47\textwidth]{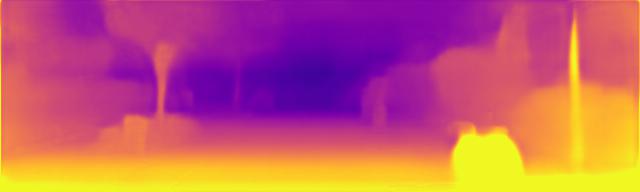}}	
	\subfloat[$\lambda_{rep} = 10^5$]{
		\includegraphics[width=0.47\textwidth]{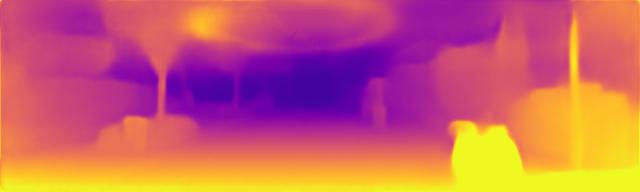}}
	\\ \vspace{-0.2cm}
	\subfloat[$\lambda_{rep} = 10^6$]{
		\includegraphics[width=0.47\textwidth]{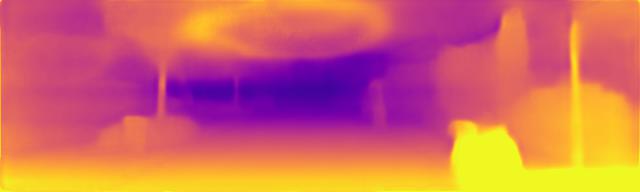}} 
    \subfloat[$\lambda_{rep} = 10^7$]{
		\includegraphics[width=0.47\textwidth]{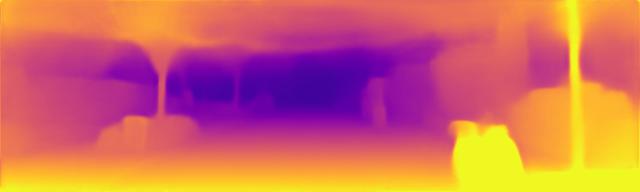}}
	\caption{\textbf{Effects of varying the coefficient} $\lambda_{rep}$ that weights the supervised loss term, for the KITTI dataset. Most noticeably, lower values of $\lambda_{rep}$ produce a better reconstruction of areas not observed by the LiDAR sensor.} 
\label{fig:change_lambda}
\end{figure*}

Interestingly, we did not notice any meaningful changes in numerical results when $\lambda_{rep}$ varies, even if this variation is by a few orders of magnitude. However, there was a significant difference in how the resulting depth maps are visually represented, as depicted in Fig. \ref{fig:change_lambda}. In particular, larger values for $\lambda_{rep}$ promote a worse reconstruction of areas not observed by the LiDAR sensor. We suspect that this behavior is due to the supervised term of the loss overwhelming the self-supervised terms, which hinders the learning of denser, smoother depth maps via the photometric loss. This is supported by the fact that this is a typical behavior of purely supervised depth learning algorithms, where the loss is never calculated in areas where there are no valid depth values. When further lowering $\lambda_{rep}$, we started to see degradation in numerical results, indicating that the photometric loss was being over-represented in the loss and scale was not being learned properly, which led us to elect $\lambda_{rep}=10^4$ as the optimal value for our proposed semi-supervised loss.

\section{Degradation in the Number of Supervised Frames}
In this section, we provide analysis of our model robustness to another type of degradation in supervision: the number of depth labels available. This is particularly useful as a way to combine large unlabeled datasets, produced without any sort of supervision, with a small amount of labeled images, obtained separately under more controlled circumstances. Our training schedule, on the KITTI dataset, consists of producing two separate splits:
\begin{itemize}
\item \textbf{Unlabeled ($\mathcal{U}$)}: All available images (39810, following the pre-processing steps of \cite{zhou2017unsupervised}) are maintained, discarding all depth information.
\item \textbf{Supervised ($\mathcal{S}$)}: $N$ images are randomly selected from the entire dataset and maintained, alongside their corresponding \textit{Annotated} depth information.
\end{itemize}

Afterwards, training is performed as instructed, however at each step half the batch size is sampled from $\mathcal{U}$ and half from $\mathcal{S}$, with the former not contributing for the proposed reprojected distance loss $\mathcal{L}_{rep}$ during loss calculation. Note that $\mathcal{S}$ is sampled with replacement, so the same labeled images can be processed multiple times in the same epoch, that is considered finished when all images from $\mathcal{U}$ are processed once. This is done to avoid data imbalance, as the number of training frames from $\mathcal{S}$ decrease relatively to $\mathcal{U}$. 

Results obtained using this training schedule are shown in Table \ref{tab:ablation}, indicating that our proposed method statistically did not degrade when observing only $10000$ images, roughly $25\%$ of the total of annotated depth maps. Additionally, when observing only $1000$ images, or $2.5\%$ the total number of annotated depth maps, our proposed methods achieved performance comparable to \citet{SemiLeftRight} and \citet{luo2018single}, considered the current state-of-the-art for semi-supervised monocular depth estimation. As we further decrease the number of supervised frames, performance starts to degrade more steeply, however these are mostly due to the model's inability to learn proper scale with such sparse (and possibly biased) information. 

\begin{figure}[b!]
    \vspace{-1.0cm}
    \centering
	\subfloat[Absolute Relative Error ($abs\_rel$).]{
		\includegraphics[width=0.45\textwidth]{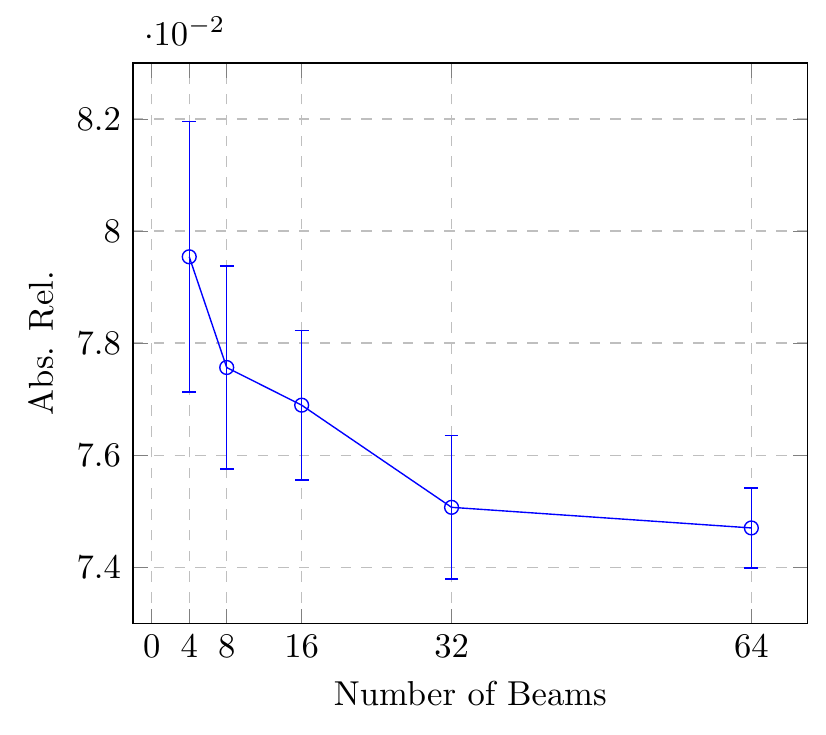}}
	\subfloat[Threshold 1 ($\delta < 1.25$)]{
		\includegraphics[width=0.45\textwidth]{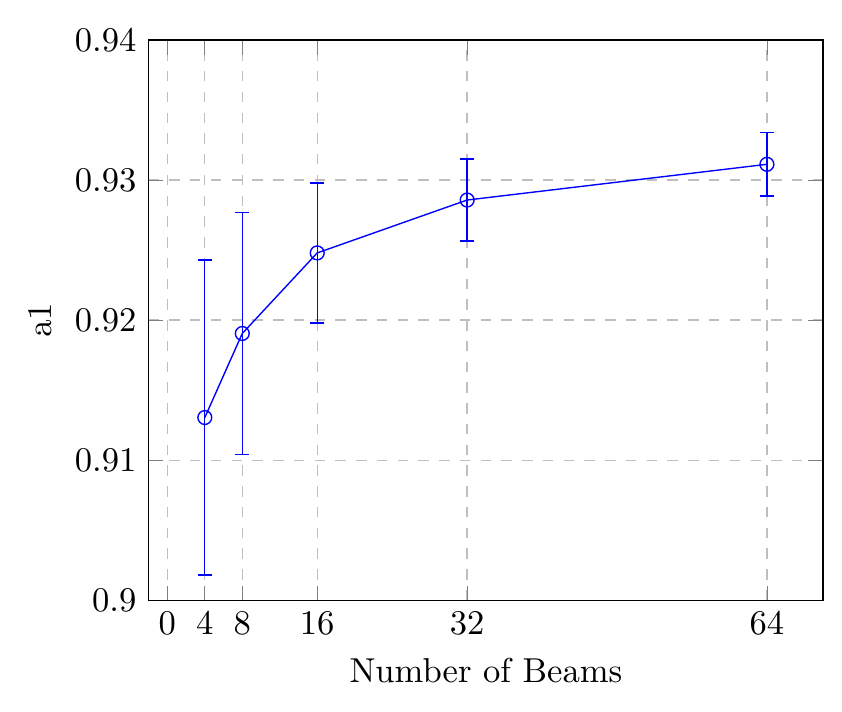}} \\
    \caption{\textbf{Effects of beam selection} in monocular depth estimation performance, for different beam distributions. The error bars indicate the variation in depth estimates when different offset values for the top beam are considered. For 64 beams, since there is no variation, the error bars are indicative of the noise inherent to stochastic training and random data augmentation.}
    \label{fig:beamsoffset}
\end{figure}

\section{Effects of Beam Selection for Sparse Depth Labels}

In this section we explore how sensitive our semi-supervised depth estimates are to the selection of beams at training time, particularly as depth labels become sparser. In other words, we would like to investigate how the distribution of valid depth pixels throughout annotated labels impact overall results. In our original experiments, beam sparsification was achieved by keeping only those at equally spaced intervals, and by increasing these intervals the number of beams decreases. Naturally, when all 64 beams are used there is no interval, when 32 are used every second beam is kept, when 16 are used every fourth beam is kept, and so forth. It is important to note that not all beams are necessarily used by the reprojected depth map, since their point of contact might not be visible by the camera. In fact, we noticed that most of the information contained in beams below the 45$^{th}$ is discarded, which makes the task of sparse semi-supervision even more challenging.

\begin{table*}[t]
\label{tab:ablation_decrease_frame}
	\centering
	{
		\small
		\setlength{\tabcolsep}{0.3em}
		\begin{tabular}{c|ccccc}
			\toprule
			
			\textbf{\# Sup. Frames} & 
			Abs.Rel &
			Sq.Rel &
			RMSE &
			RMSE$_{log}$ &
			$\delta < 1.25$
			\vspace{0.5mm}\\
			\toprule


\textbf{39810 (all)} 
& 0.073 $\pm$ 0.001
& 0.344 $\pm$ 0.004
& 3.273 $\pm$ 0.008
& 0.117 $\pm$ 0.001
& 0.932 $\pm$ 0.002

\\


\textbf{10000}  
& 0.074 $\pm$ 0.002
& 0.346 $\pm$ 0.006
& 3.298 $\pm$ 0.021
& 0.118 $\pm$ 0.002
& 0.934 $\pm$ 0.002

\\


\textbf{1000}  
& 0.080 $\pm$ 0.003
& 0.388 $\pm$ 0.010
& 3.550 $\pm$ 0.038
& 0.125 $\pm$ 0.005
& 0.923 $\pm$ 0.004 

\\


\textbf{100}  
& 0.101 $\pm$ 0.007
& 0.532 $\pm$ 0.023
& 4.230 $\pm$ 0.078
& 0.155 $\pm$ 0.018
& 0.886 $\pm$ 0.013

\\


\textbf{10}  
& 0.249 $\pm$ 0.031
& 2.832 $\pm$ 0.081
& 10.412 $\pm$ 0.380
& 0.439 $\pm$ 0.059
& 0.561 $\pm$ 0.047

\\

\bottomrule
\end{tabular}
	}\vspace{1mm}
	\caption{\textbf{Quantitative results} showing how our proposed semi-supervised methodology behaves with a decreasing number of supervised frames at training time, for the KITTI dataset. For each row, statistical intervals were calculated based on 10 independent models trained using different random subsets from $\mathcal{S}$. For \textbf{all}, the entire $\mathcal{S}$ was used in all 10 sessions, with the statistical intervals being indicative of the noise inherent to stochastic training and random data augmentation.}

	\label{tab:ablation}
\end{table*}

In order to vary the position of depth information in the resulting sparse labels, while maintaining a proper distribution similar to what a real LiDAR sensor would provide, we opted for introducing an offset, determining where the top beam is located. Starting from 0, this offset increases until it coincides with another beam that was selected when no offset is considered. Following this strategy, when 32 beams are considered there are 2 variations, when 16 beams are considered there are 4, and so forth. The results when using this strategy are depicted in Fig. \ref{fig:beamsoffset}, where we can see that sparser depth labels are more sensitive to the distribution of valid pixels, and there are indeed some configurations that lead to better results, however there was no configuration that resulted in catastrophic failures. Interestingly, as we further increased sparsity, considering only 2 or even 1 beam, some configurations failed to converge, showing that there is a limit to how much sparsity  can be properly leveraged in our proposed semi-supervised learning framework, however a more thorough analysis is left for future work.

\begin{figure}[t!]
    \vspace{-0.5cm}
	\centering
	\subfloat[\textit{Annotated} depth maps]{
		\includegraphics[height=2.7cm,width=0.32\textwidth]{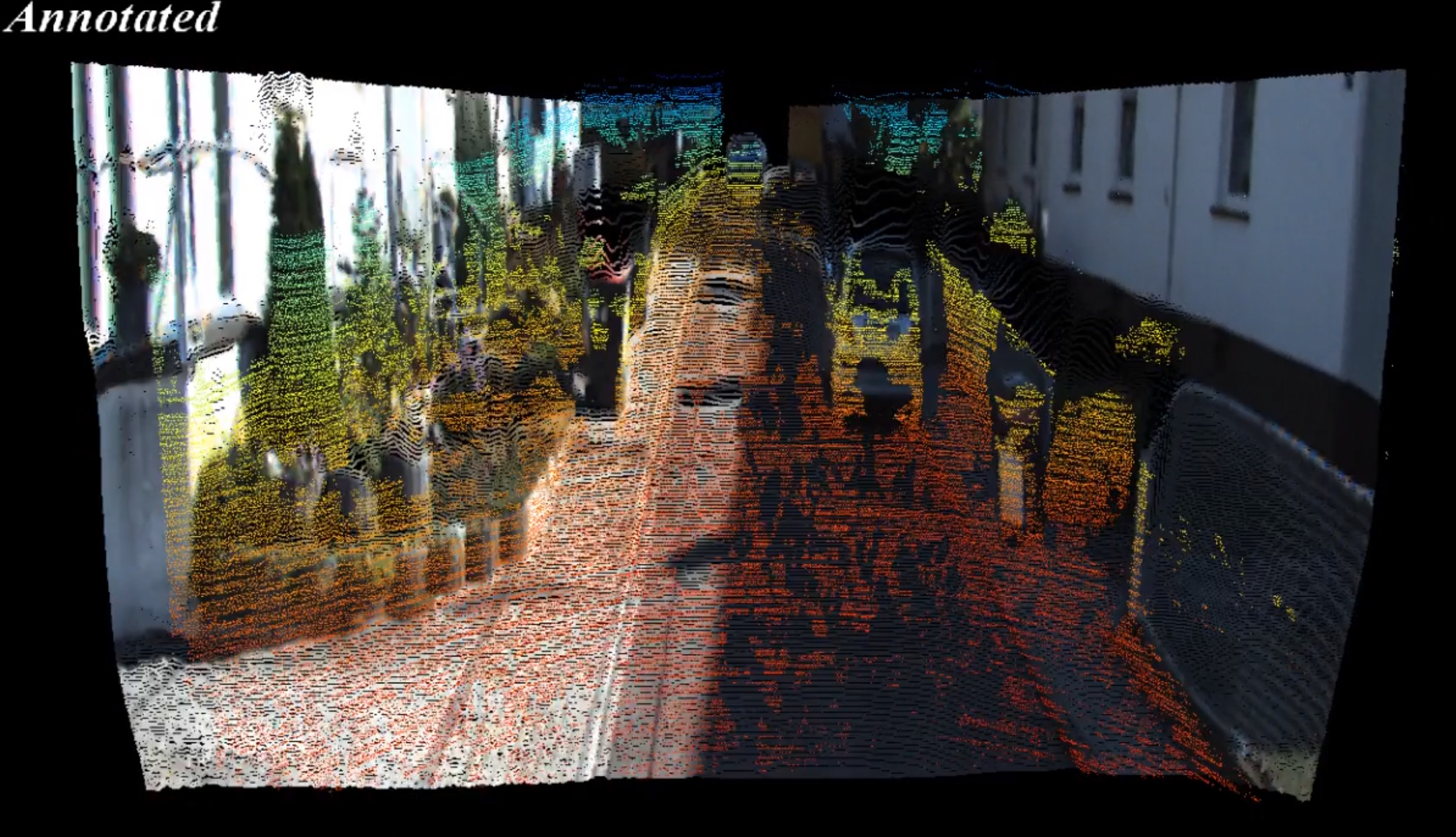}}	
	\subfloat[64 beams]{
		\includegraphics[height=2.7cm,width=0.32\textwidth]{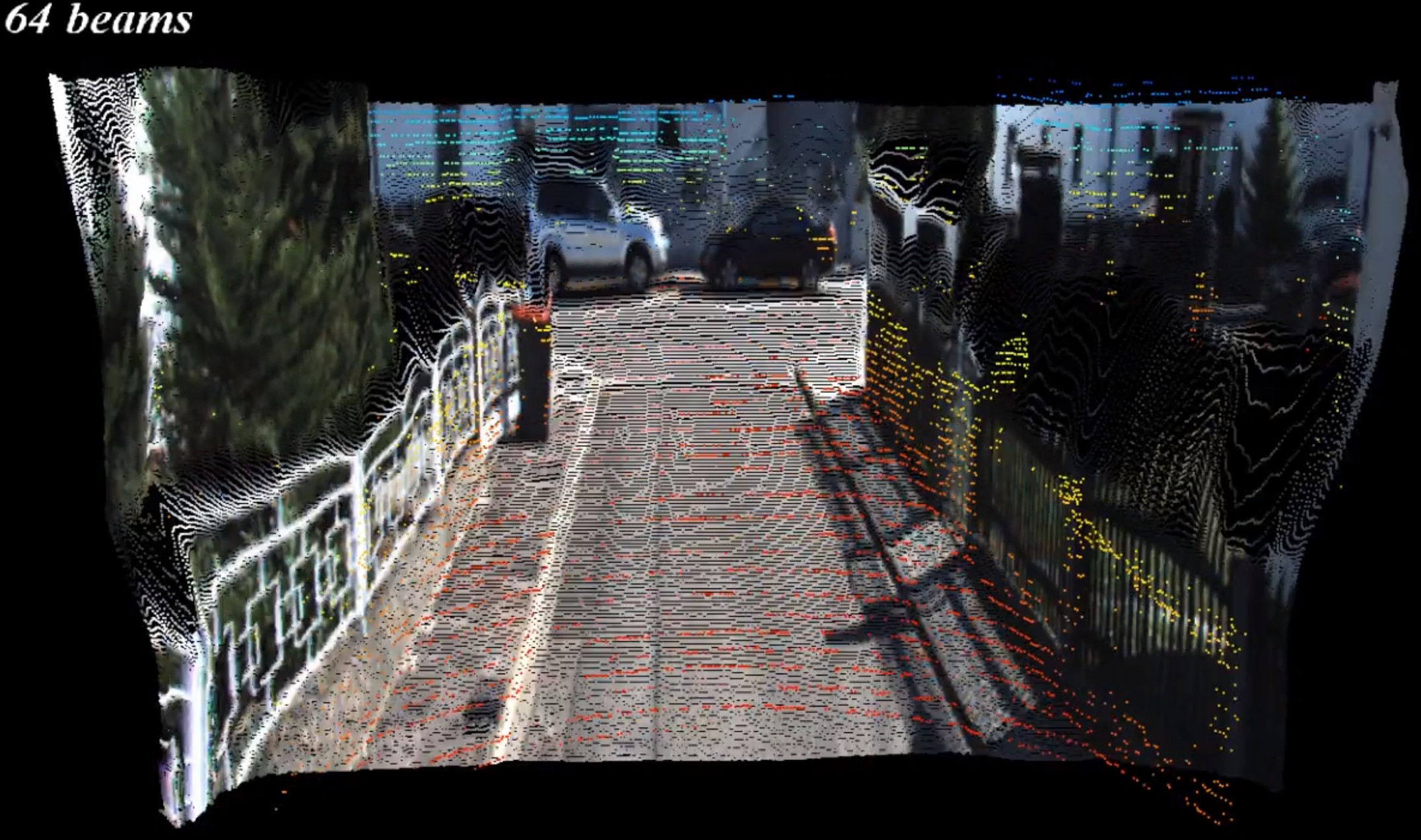}}
	\subfloat[32 beams]{
		\includegraphics[height=2.7cm,width=0.32\textwidth]{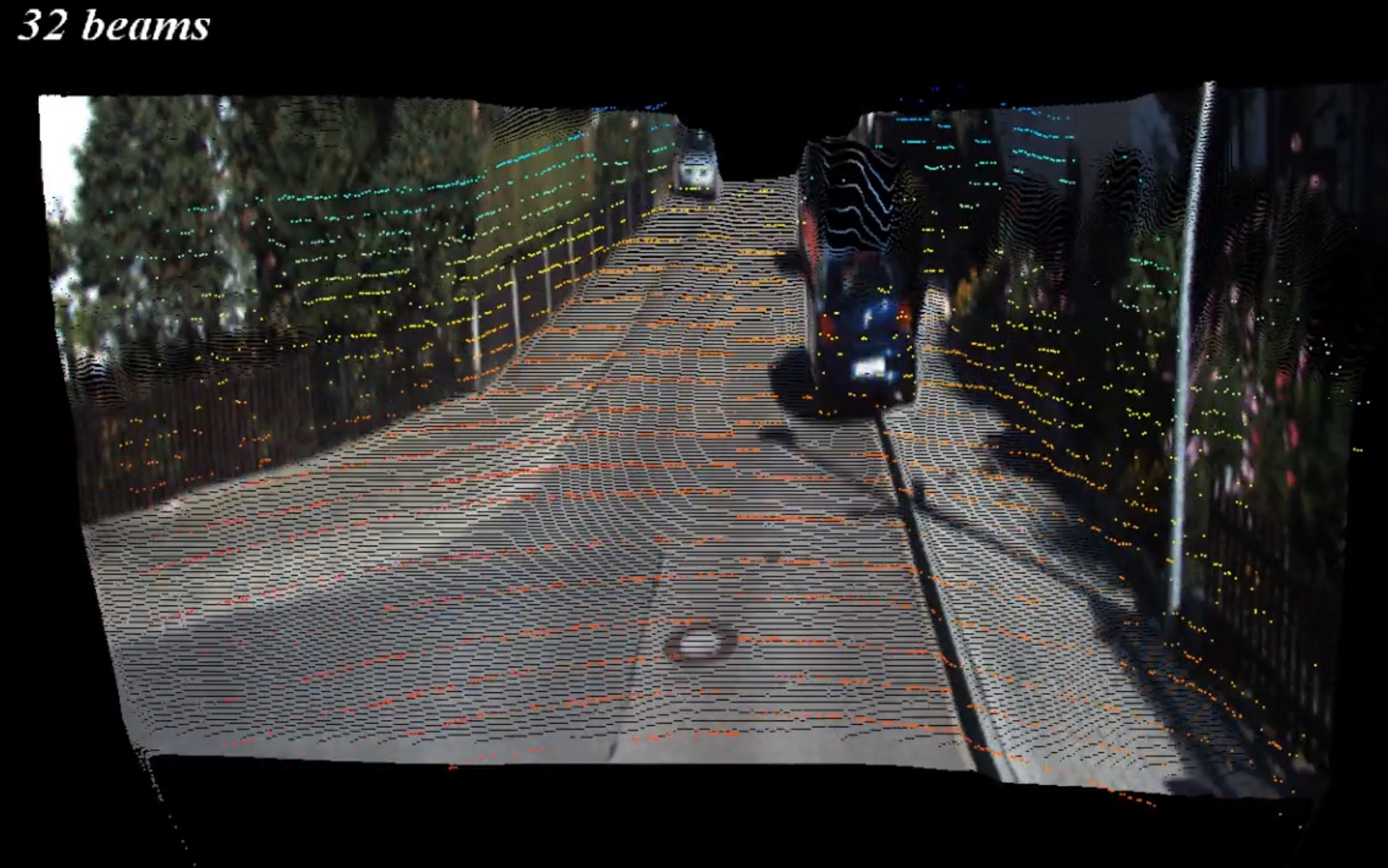}}
	\\\vspace{-0.3cm}
    \subfloat[16 beams]{
		\includegraphics[height=2.7cm,width=0.32\textwidth]{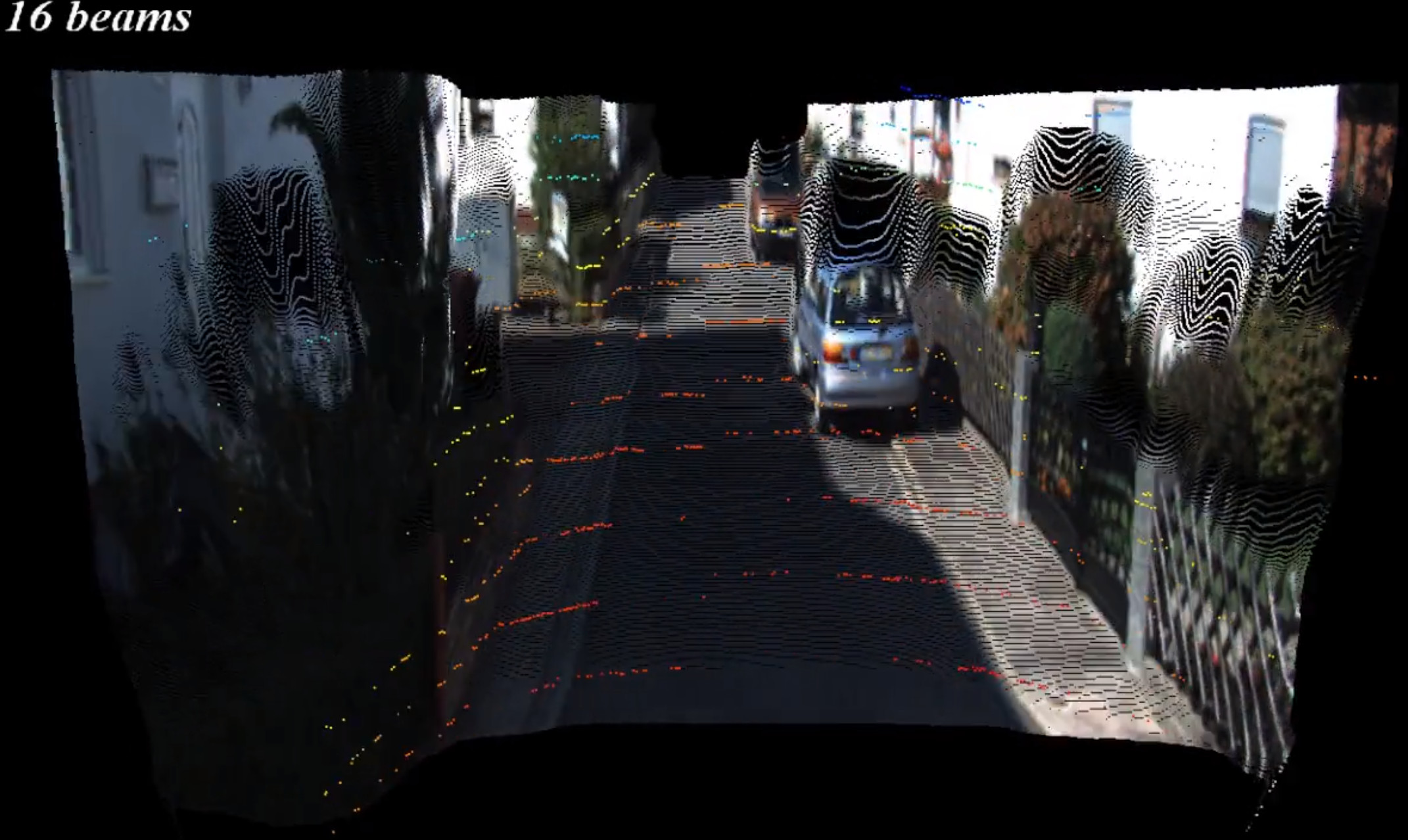}}
    \subfloat[8 beams]{
		\includegraphics[height=2.7cm,width=0.32\textwidth]{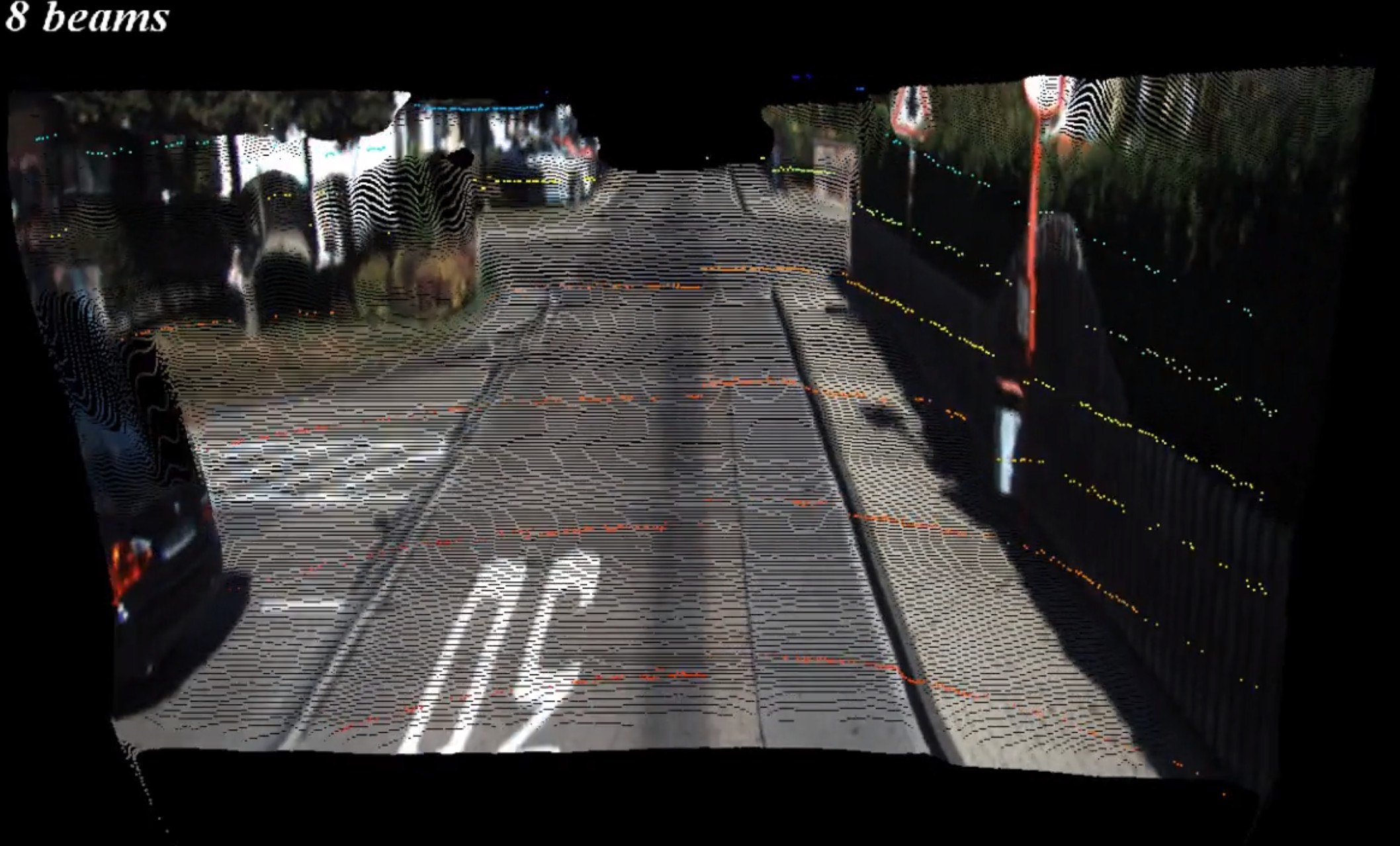}}
    \subfloat[4 beams]{
		\includegraphics[height=2.7cm,width=0.32\textwidth]{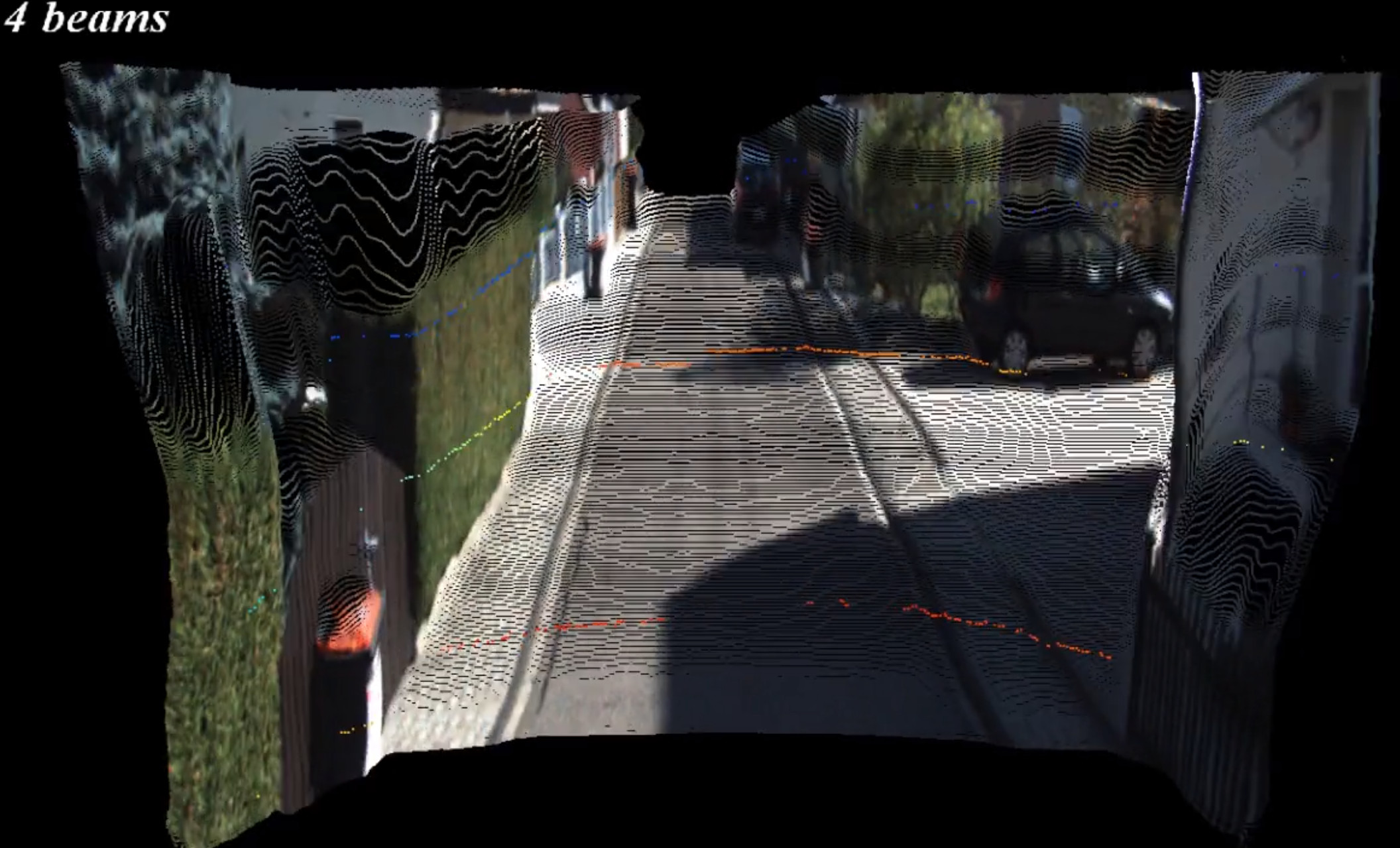}}
	\caption{\textbf{Reconstructed point-clouds} from our proposed semi-supervised depth estimation methodology, with models trained using different numbers of LiDAR beams.} 
\label{fig:videobeams}
\end{figure}

\section{Additional Qualitative Results}
\begin{figure}[t!]
    \vspace{-0.8cm}
	\centering
	\subfloat{
		\includegraphics[height=1.95cm,width=0.50\textwidth]{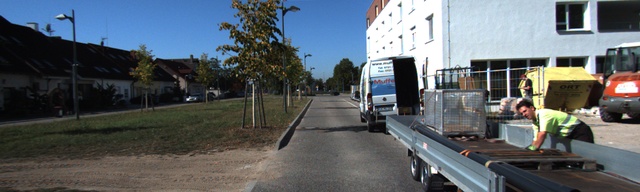}}     
    \subfloat{
		\includegraphics[height=1.95cm,width=0.50\textwidth]{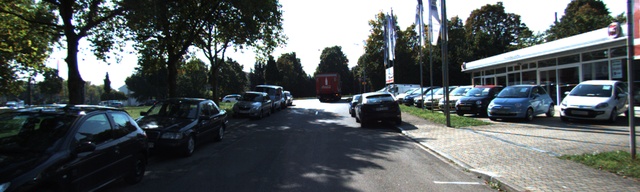}}
\\ \vspace{-0.38cm}
	\subfloat{
		\includegraphics[height=1.95cm,width=0.50\textwidth]{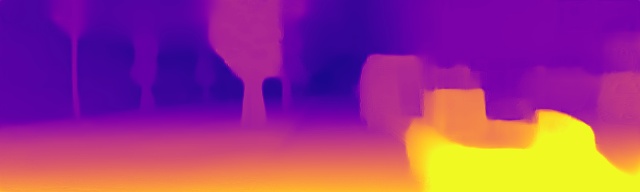}}     
    \subfloat{
		\includegraphics[height=1.95cm,width=0.50\textwidth]{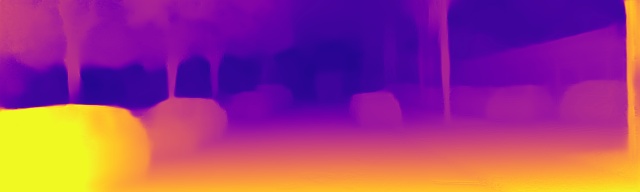}}
\\ \vspace{-0.3cm}
    \subfloat{
	    \includegraphics[height=1.95cm,width=0.50\textwidth]{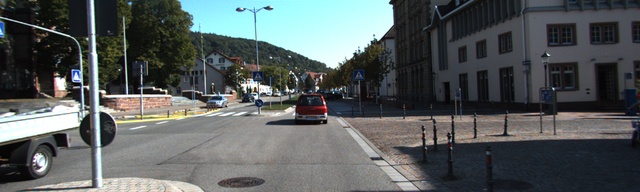}}		
    \subfloat{
		\includegraphics[height=1.95cm,width=0.50\textwidth]{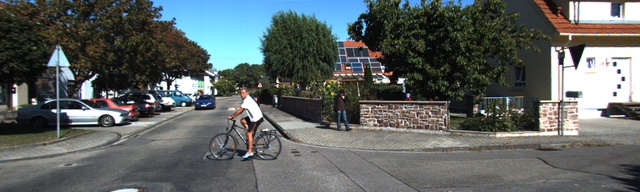}}
\\ \vspace{-0.38cm}
	\subfloat{
		\includegraphics[height=1.95cm,width=0.50\textwidth]{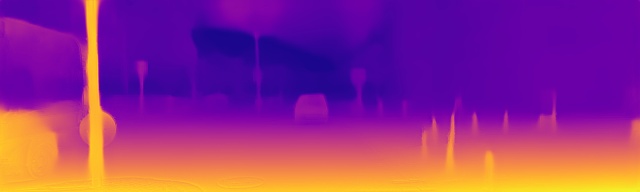}}     
    \subfloat{
		\includegraphics[height=1.95cm,width=0.50\textwidth]{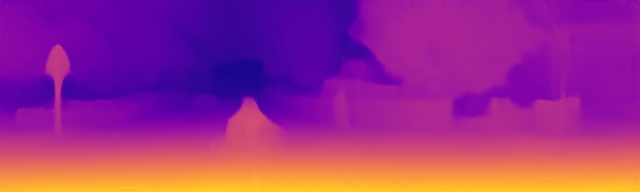}}
\\ \vspace{-0.3cm}
	\subfloat{
		\includegraphics[height=1.95cm,width=0.50\textwidth]{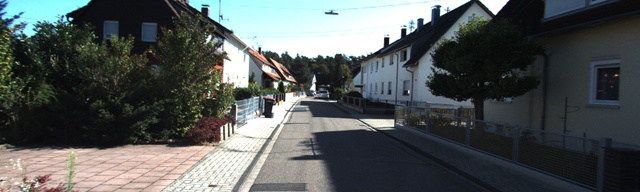}}     
    \subfloat{
		\includegraphics[height=1.95cm,width=0.50\textwidth]{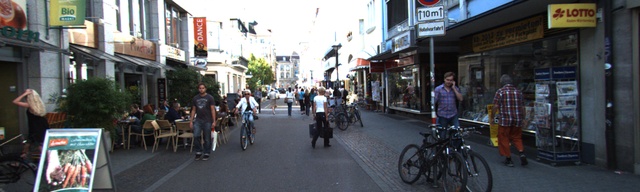}}
\\ \vspace{-0.38cm}
	\subfloat{
		\includegraphics[height=1.95cm,width=0.50\textwidth]{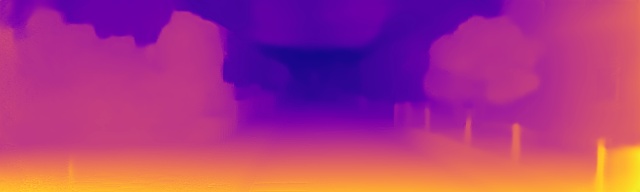}}     
    \subfloat{
		\includegraphics[height=1.95cm,width=0.50\textwidth]{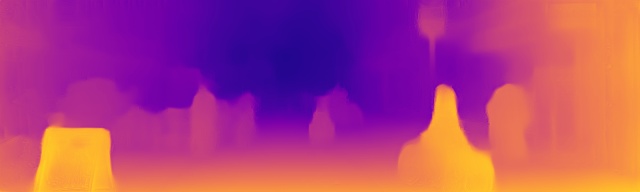}}
\\ \vspace{-0.3cm}		
    \subfloat{
		\includegraphics[height=1.95cm,width=0.50\textwidth]{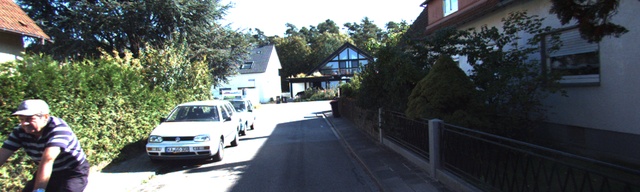}} 
    \subfloat{
		\includegraphics[height=1.95cm,width=0.50\textwidth]{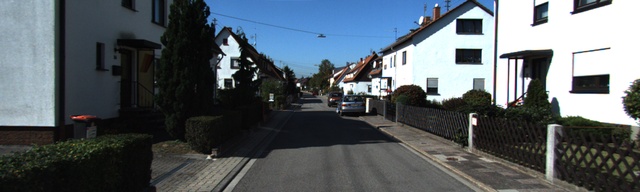}}		
\\ \vspace{-0.38cm}
	\subfloat{
		\includegraphics[height=1.95cm,width=0.50\textwidth]{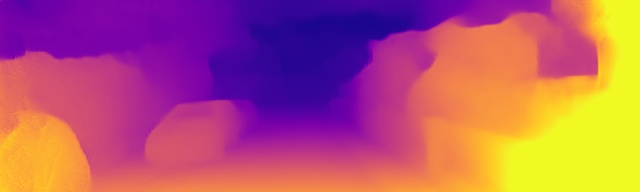}}     
    \subfloat{
		\includegraphics[height=1.95cm,width=0.50\textwidth]{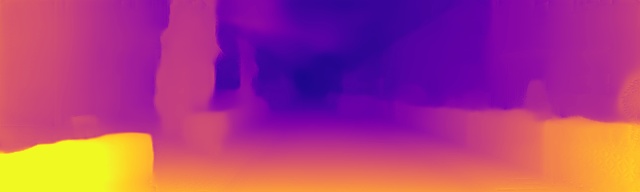}}
\\ \vspace{-0.3cm}
	\subfloat{
		\includegraphics[height=1.95cm,width=0.50\textwidth]{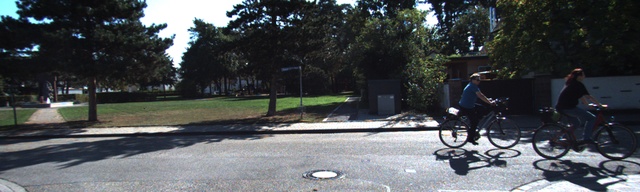}}     
    \subfloat{
		\includegraphics[height=1.95cm,width=0.50\textwidth]{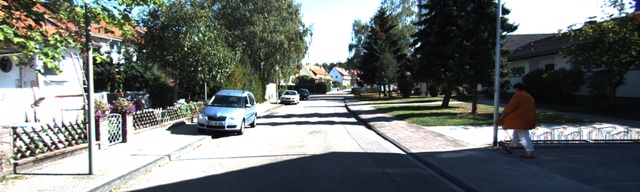}}
\\ \vspace{-0.38cm}
    \subfloat{
		\includegraphics[height=1.95cm,width=0.50\textwidth]{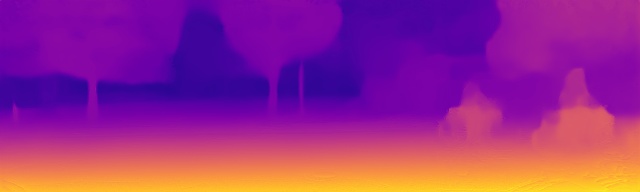}} 
    \subfloat{
		\includegraphics[height=1.95cm,width=0.50\textwidth]{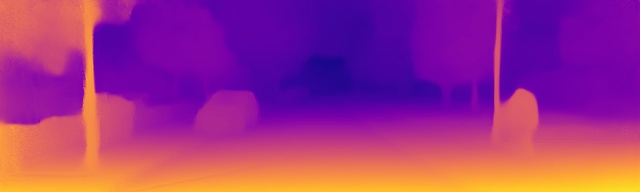}} 
	\caption{\textbf{Qualitative results} of our proposed semi-supervised monocular depth estimation methodology, showing input RGB images and output depth maps.}
\label{fig:supqualitative}
\end{figure}

Here we provide some more qualitative results of our proposed semi-supervised monocular depth estimation methodology, using the reprojected distance loss, on the KITTI dataset. Fig. \ref{fig:videobeams} depicts reconstructed pointclouds from models trained using different numbers of LiDAR beams, while Fig. \ref{fig:supqualitative} shows corresponding input RGB images and output depth maps. More qualitative results can be found on the supplementary video attached.

\bibliography{references}  
\clearpage

\end{document}